\documentclass[letterpaper, 10 pt, conference]{ieeeconf}  

\IEEEoverridecommandlockouts                              

\usepackage{cite}
\usepackage[pdftex]{graphicx}
\graphicspath{{figs/}}
\usepackage{amsmath}
\usepackage{amsfonts}
\usepackage{comment}
\usepackage{algorithmic}
\usepackage[ruled,vlined,linesnumbered]{algorithm2e}
\usepackage[dvipsnames]{xcolor}
\usepackage{mathtools}
\usepackage[hidelinks]{hyperref}
\usepackage{multirow}
\usepackage{booktabs}
\usepackage{siunitx}

\usepackage{balance}

\usepackage{stmaryrd}

\usepackage{float}

\usepackage{cuted}

\usepackage{placeins}

\usepackage{dblfloatfix}

\newcommand{\citep}[1]{\cite{#1}}

\newcounter{equationset}

\title{\LARGE \bf
Transitional Grid Maps: \\Joint Modeling of Static and Dynamic Occupancy
}

\author{José Manuel Gaspar Sánchez$^{1}$, Leonard Bruns$^{2}$, Jana Tumova$^{2}$, Patric Jensfelt$^{2}$ and Martin Törngren$^{1}$
\thanks{*This research has been carried out as part of the TECoSA Vinnova Competence Center for Trustworthy Edge Computing Systems and Applications and partly supported through Entice (Vinnova, project no. 2022-03000).}
\thanks{$^{1}$José Manuel Gaspar Sánchez and Martin Törngren are with the Mechatronics Division, KTH Royal Institute of Technology, Stockholm, Sweden (jmgs@kth.se; martint@kth.se).}%
\thanks{$^{2}$Leonard Bruns, Jana Tumova and Patric Jensfelt are with the Division of Robotics, Perception and Learning, KTH Royal Institute of Technology, Stockholm, Sweden (leonardb@kth.se; tumova@kth.se; patric@kth.se).}%
\thanks{We thank Eric Gil Alonso for his help during the data collection.}%
}

\begin{document}

\thinmuskip=1mu
\medmuskip=1mu
\thickmuskip=1mu

\maketitle
\thispagestyle{empty}
\pagestyle{empty}

\begin{abstract}

Autonomous agents rely on sensor data to construct representations of their environments, essential for predicting future events and planning their actions. However, sensor measurements suffer from limited range, occlusions, and sensor noise. These challenges become more evident in highly dynamic environments.
This work proposes a probabilistic framework to jointly infer which parts of an environment are statically and which parts are dynamically occupied.
We formulate the problem as a Bayesian network and introduce minimal assumptions that significantly reduce the complexity of the problem. Based on those, we derive Transitional Grid Maps (TGMs), an efficient analytical solution.
Using real data, we demonstrate how this approach produces better maps by keeping track of both static and dynamic elements and, as a side effect, can help improve existing SLAM algorithms.
\end{abstract}

\section{Introduction}

The environments populated by autonomous agents include two components: the static and the dynamic part. Consider for example an automated vehicle, which has to identify dynamic obstacles, such as other vehicles and pedestrians, and static elements, such as buildings.
The static part is often used for localization and global path planning. 
In contrast, the dynamic part of the environment, i.e., the set of other traffic participants, is typically used for local planning since the ego vehicle needs to react to it.

A common approach to representing these environments is to decouple both components, i.e., to use specialized methods for the static and the dynamic parts independently. For example, an occupancy grid map (OGM) \citep{elfes_using_1989, moravec_high_1985, moravec_sensor_1989} can be used to create a map of the static environment while a multi-target tracker \cite{wang2007simultaneous} can be used to maintaining a list of known dynamic obstacles and track them independently.

A challenge with this approach is that each method assumes to get inputs only from the part of the environment they are designed for. Grid maps assume the environment is completely static, and object trackers assume to only get observations from dynamic obstacles. For this approach to work, raw sensor measurements must be classified into static and dynamic. In some cases, this can be done using learning-based semantic segmentation. Unfortunately, this approach becomes less reliable for sparse or noisy sensors, such as radars and low-resolution lidars, as well as for difficult environmental conditions such as heavy rain or snow.

An alternative is to use methods able to both map the static environment and track dynamic obstacles simultaneously. Some of the approaches developed in this area have modeled the environment as cells, similar to the occupancy grid maps, including also a velocity distribution for each cell, and modeling the temporal and spatial relations between cells as a Bayesian network, as in \citep{coue_using_2003, coue_bayesian_2006}.
Unfortunately, in the general case, probabilistic inference in Bayesian networks is NP-hard \citep{cooper_computational_1990}, which makes these approaches computationally intractable unless they introduce significant simplifications, such as only updating the portion of the map in the current field of view of the agent.

\begin{figure}[t]
\vspace{0.2cm}
    \centering
    \includegraphics[width=.8\linewidth]{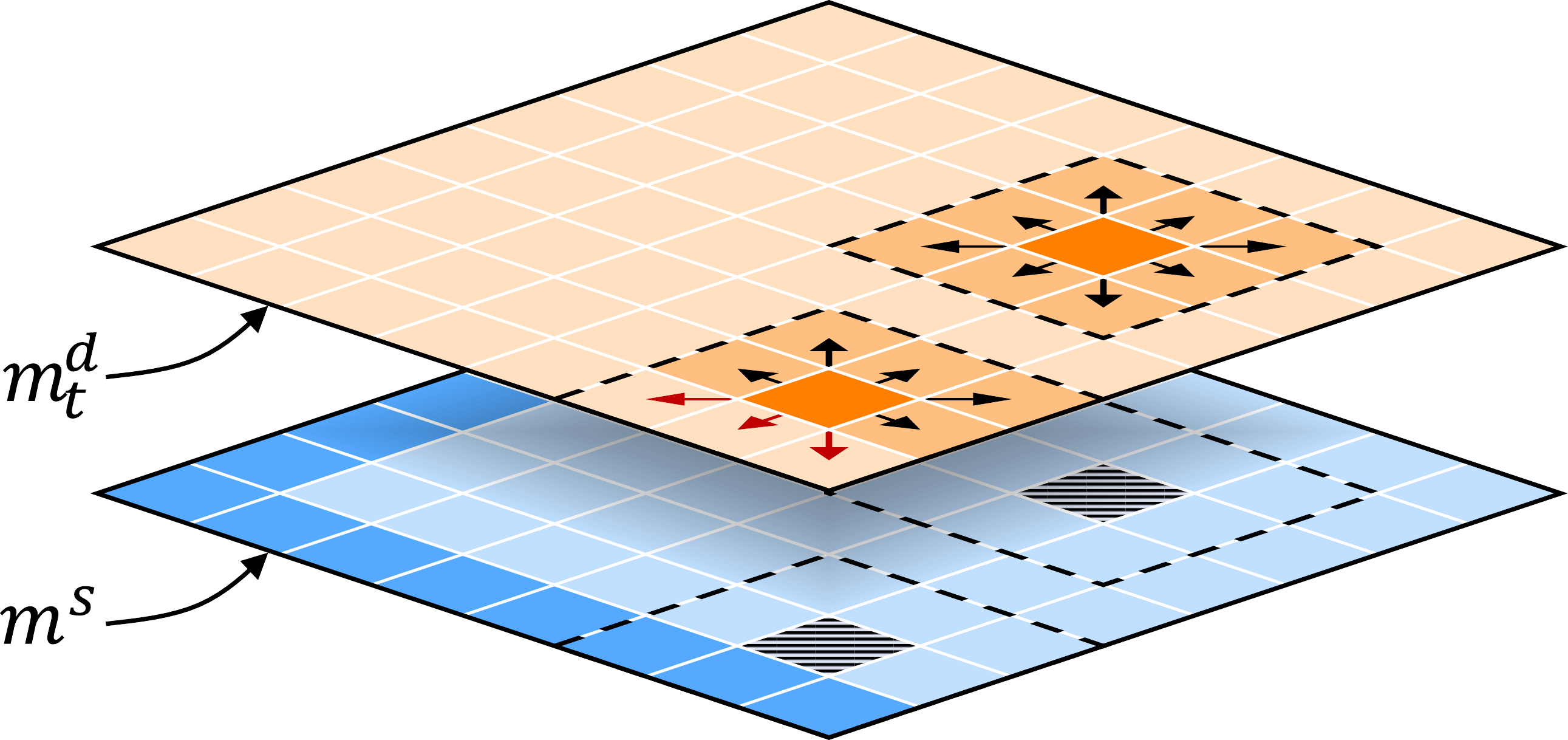}
    \caption{Transitional Grid Map (TGM). The dynamic environment at the current time, $m^d_t$, is predicted based on the transition probability between cells, represented with arrows, and the current belief about the static environment, $m^s$, which constrains the feasible transitions.}
    \vspace{-0.5cm}
    \label{fig:intro}
\end{figure}

In this work, we model the environment as two components: static and dynamic, as represented in Fig.~\ref{fig:intro}.
Measurements are subject to noise and affected by both components but it is not directly possible to decouple the influence of each component on the individual measurements.
We formulate the problem of inferring the static and dynamic components of an environment as a Bayesian network, introducing a set of assumptions to make the computation tractable.

\subsection{Contributions}
This work studies \emph{the inference of the static and dynamic parts of the environment with limited sensing}.
The main contributions are:

\begin{enumerate}
    \item a mathematical formulation of the problem of inferring which parts of an environment are statically and which parts are dynamically occupied, analyzing where the computation becomes intractable;
    \item formulating assumptions that hold in many practical scenarios and allow for the problem to be simplified, deriving Transitional Grid Maps (TGMs) as an efficient analytical solution;
    \item an experimental evaluation of the resulting TGMs, showing how they produce more accurate maps and, as a side effect, benefit existing SLAM algorithms.
\end{enumerate}
In the following section, we elaborate how these contributions advance the state of the art. 

\subsection{Related Work}
\label{sec:related_work}

Occupancy grid maps (OGMs), \citep{elfes_using_1989, moravec_high_1985, moravec_sensor_1989}, are used to fuse multiple sensor measurements into a unified representation of the environment. The environment is modeled as a set of cells, each of which has a binary state, occupied or free. Assuming a static environment and independence between the cells, the probability of occupancy of each cell of the map can be computed individually.

However, the real world is often not static. When it is not, the original formulation of the traditional grid map method trusts previous beliefs over new contradicting measurements; it takes the same number of observations to change the state of a cell as past observations that were used to set it \citep{yguel_update_2008}.

Some modifications have been suggested to address this, such as limiting how confident the method becomes, as in \citep{kraetzschmar_probabilistic_2004, hornung2013octomap}, or penalizing old information with a decay factor \citep{yguel_update_2008}.
These methods suffer from the fact that the underlying model is still derived assuming a static environment and, to compensate for this, they penalize all the accumulated knowledge equally. Instead, we model two kinds of occupied cells, static and dynamic, and infer which class they likely belong to based on the measurements over time.

The method proposed in \citep{meyer-delius_occupancy_2012} uses a hidden Markov model (HMM) for each cell to represent both the belief about the state of the cell and the corresponding probability of state change. While this approach explicitly models how occupancy changes over time, each cell is represented individually, which ignores that obstacles cannot suddenly appear but must have traveled there from another neighboring cell. Also employing an HMM, \cite{wang_modeling_2014} additionally incorporates observations from neighboring cells at the previous time step when estimating the state of each cell.
However, when no observations in neighboring cells are available, e.g., during occlusions, their model only relies on the individual cell state.
We instead assume a transition model between cells, which makes cells dependent on the belief about the previous state of neighboring cells, even the ones currently occluded.

Other authors have used particle filters to represent dynamic obstacles \citep{steyer_object_2017, steyer_grid-based_2018, tanzmeister_evidential_2017,nuss_random_2018}.
Most of them make use of the Dempster-Shafer theory of evidence \citep{dempster_generalization_1968, shafer_mathematical_1976}, which allows to only populate cells with particles where there is some evidence of occupancy. This approach, used to reduce computation, implies that areas outside the current field of view are not populated with particles.

A pure Bayesian particle filter, without the Dempster-Shafer theory, is presented in \citep{rexin_modeling_2017} to keep track of occluded areas. This filter also populates particles in occluded areas to represent the hypotheses of possible unseen obstacles, leading to a drastic increase in the number of particles generated. Computation times are not analyzed in this work, but because of the number of particles required, it may not be suitable for a real-time implementation.

More recently, learning-based methods operating on grids have been proposed to perceive and predict dynamic obstacles \cite{dequaire_deep_2018, schreiber_dynamic_2021}. These approaches, while showing promising results, still require extensive training and their performance depends on the quality of the available data.
\setcounter{equation}{1}
\begin{table*}[b]
\centering
\label{tab:1}
\begin{minipage}{\textwidth}
\hrule
\begin{align}
    p(m_{t,i} \mid z_{1:t}, x_{1:t}) &= \frac{p(z_t \mid m_{t,i}, z_{1:t-1}, x_{1:t})p(m_{t,i} \mid z_{1:t-1}, x_{1:t})}{p(z_t \mid z_{1:t-1}, x_{1:t})}
    && \textit{Bayes rule} \label{eq:derivation1.1}\\
    &= \frac{p(z_t \mid m_{t,i}, z_{1:t-1}, x_{1:t})p(m_{t,i} \mid z_{1:t-1}, \textcolor{blue}{x_{1:t-1}})}{p(z_t \mid z_{1:t-1}, x_{1:t})}
    && \textit{Independence} \label{eq:derivation1.2}\\
    &\approx \frac{\textcolor{blue}{p(z_t \mid m_{t,i}, x_t)} p(m_{t,i} \mid z_{1:t-1}, x_{1:t-1})}{p(z_t \mid z_{1:t-1}, x_{1:t})}
    && \textit{Markov assumption} \label{eq:derivation1.3}\\
    &= \frac{\textcolor{blue}{p(m_{t,i} \mid z_t, x_t) p(z_t \mid x_t)} p(m_{t,i} \mid z_{1:t-1}, x_{1:t-1})}{\textcolor{blue}{p(m_{t,i} \mid x_t)} p(z_t \mid z_{1:t-1}, x_{1:t})}
    && \textit{Bayes rule} \label{eq:derivation1.6}\\
    &= \frac{p(m_{t,i} \mid z_t, x_t) p(z_t \mid x_t) p(m_{t,i} \mid z_{1:t-1}, x_{1:t-1})}{\textcolor{blue}{p(m_{t,i})} p(z_t \mid z_{1:t-1}, x_{1:t})}
    && \textit{Independence} \label{eq:derivation1.7}\\
    &= \textcolor{blue}{\mu} \frac{p(m_{t,i} \mid z_t, x_t) p(m_{t,i} \mid z_{1:t-1}, x_{1:t-1})}{p(m_{t,i})}
    && \textit{Total probability} \label{eq:derivation1.8}
\end{align}
\medskip
\end{minipage}
\end{table*}

\section{Method}
\label{sec:method}

\subsection{Problem Formulation}
\label{sec:Problem}

We consider an agent with known poses, position and orientation, up to the current time $t$, $x_{1:t} = \{x_1, x_2, ..., x_t\}$, equipped with a range sensor, e.g., a lidar. The assumption about the known poses can be relaxed using existing SLAM algorithms as shown in Sec~\ref{sec:SLAM}. The sensor has a limited range, is limited by occlusions, and is subject to noise. It produces a set of measurements $z_{1:t} = \{z_1, z_2, ..., z_t\}$. The agent moves in an environment populated by dynamic and static obstacles. The problem at hand is to estimate the state of the environment, i.e., to estimate which parts of the environment are free, static (occupied by a static obstacle) or dynamic (occupied by a dynamic obstacle).

\subsection{Bayesian Modeling}

To address this problem, we model the environment $m_t$ at each time $t$  as a set of N cells, i.e., $m_t = \{ m_{t,1}, m_{t,2}, ..., m_{t,N} \}$. Each cell can be in one of three possible states: free, static, or dynamic; i.e., $m_{t,i} \in \{ f,s,d \}$.
For clarity of the notation, we also define the binary variables ${m^s_{t,i}=\llbracket m_{t,i}=s \rrbracket}$, ${m^d_{t,i}= \llbracket m_{t,i}=d \rrbracket}$, and ${m^f_{t,i}=\llbracket m_{t,i}=f \rrbracket}$, where $\llbracket \ \rrbracket$ denotes the Iverson bracket.

The static part of the environment, ${m^s = \{ m^s_1, m^s_2, ..., m^s_N \}}$, is assumed to stay constant, thus, it will be referred to without the time index. In contrast, the dynamic environment, ${m^d_t = \{ m^d_1, m^d_2, ..., m^d_N \}}$, can change over time, as dynamic obstacles are able to move through space. This means that the state of a dynamic cell can transition to nearby cells over time, with the restriction that they can not transition to a static cell.

\begin{figure}[htb]
    \vspace{-0.1cm}
    \centering
    \includegraphics[width=.8\linewidth]{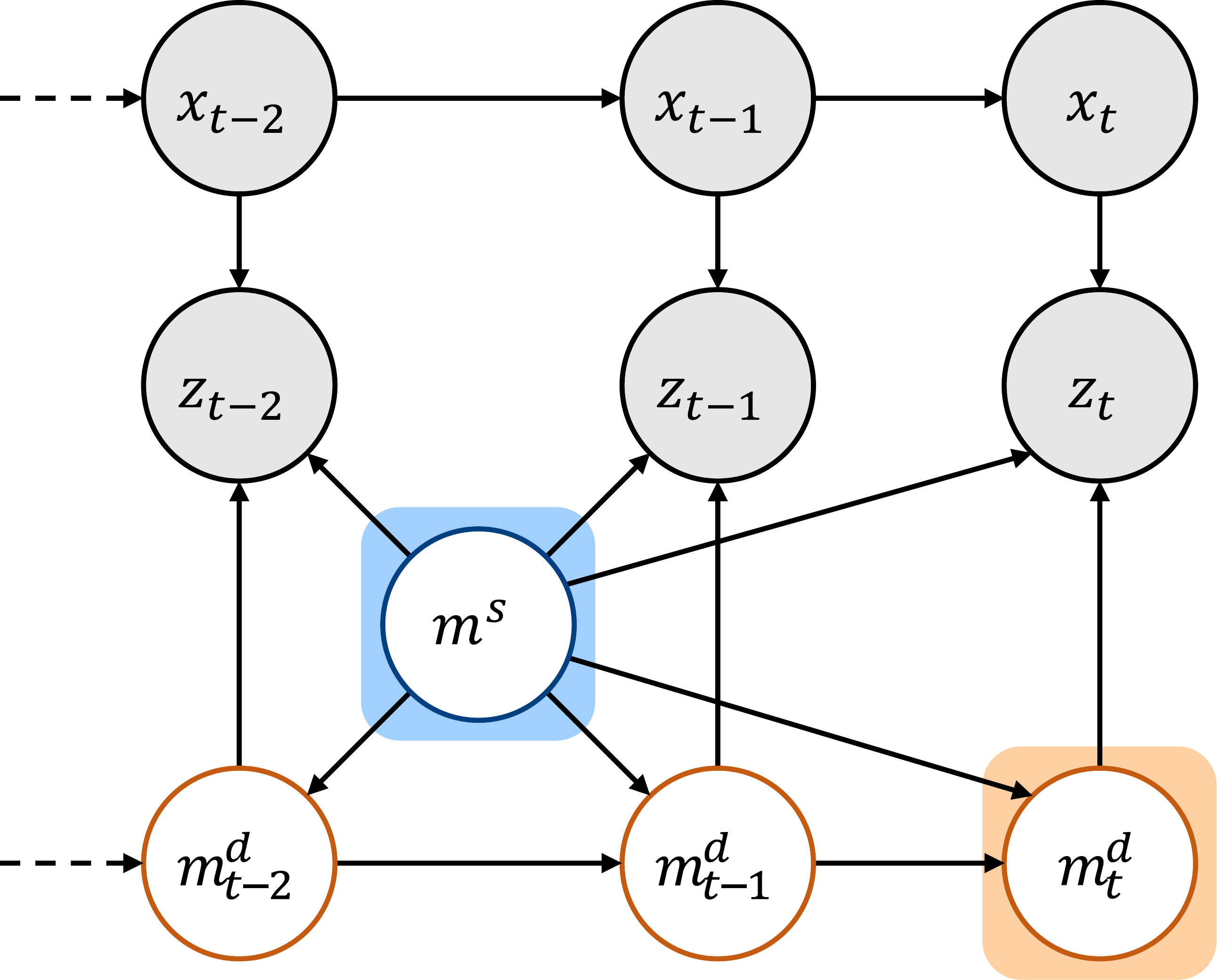}
    \caption{Bayesian network of mapping a dynamic environment with known poses. Poses $x_{1:t}$ and measurements $z_{1:t}$ are known, and the goal is to estimate the static part of the environment $m^s$ and the current state of the dynamic part of the environment $m^d_t$.}
    \label{fig:model}
\end{figure}

Figure~\ref{fig:model} shows the proposed graphical model of the problem of mapping a dynamic environment. The agent's poses $x_{1:t}$ and measurements $z_{1:t}$ are known up to the current time. The map is unknown and can be split into two parts: a static map $m^s$ that stays constant over time and a dynamic map $m^d_t$ that evolves stochastically, influenced by the static map since dynamic obstacles can not move to static cells. Measurements depend on the robot's pose, the static map and the dynamic map, but static and dynamic obstacles are assumed to be indistinguishable from a single measurement.

Starting at the same assumption as the original grid mapping algorithm for static environments, we assume that the problem can be broken down into the estimation of individual cells, such that the posterior of the map is approximated as the product of its marginals \citep{elfes_using_1989, moravec_high_1985, moravec_sensor_1989}:
\setcounter{equation}{0}
\begin{equation}
    \label{eq:marginals}
    p(m_{t} \mid z_{1:t}, x_{1:t}) \approx \prod_{i} p(m_{t,i} \mid z_{1:t}, x_{1:t}).
\end{equation}

Following the derivations in \eqref{eq:derivation1.1}-\eqref{eq:derivation1.8}, with changes \textcolor{blue}{highlighted}, the posterior probability for each possible state is
\setcounter{equation}{7}
\begin{align}
    \overbrace{p(m^s_{i} \mid z_{1:t}, x_{1:t})}^{\textit{posterior}} &= \eta \frac{\overbrace{p(m^s_{i} \mid z_t, x_t)}^{\textit{inv. sensor model}} \overbrace{p(m^s_{i} \mid z_{1:t-1}, x_{1:t-1})}^{\textit{prediction}}}{p(m^s_{i})}, \notag \\
    p(m^d_{t,i} \mid z_{1:t}, x_{1:t}) &= \eta \frac{p(m^d_{t,i} \mid z_t, x_t) p(m^d_{t,i} \mid z_{1:t-1}, x_{1:t-1})}{p(m^d_{t,i})}, \notag \\
    p(m^f_{t,i} \mid z_{1:t}, x_{1:t}) &= \eta \frac{p(m^f_{t,i} \mid z_t, x_t) p(m^f_{t,i} \mid z_{1:t-1}, x_{1:t-1})}{\underbrace{p(m^f_{t,i})}_{\textit{prior}}}.
\end{align}

The updated belief about each cell having one of the three possible states (\textit{posterior}), is computed based on the current observation (\textit{inverse sensor model}), the previous observations (\textit{prediction}), and the initial probability (\textit{prior}). To ensure the resulting products are valid probabilities, they are normalized with the normalization constant $\eta$.

The \textit{priors} are set beforehand, typically equal. The \textit{inverse sensor model} can be computed similarly to how it is done for a regular grid map, with the only distinction being that the probability of a cell being occupied has to be split between static and dynamic proportionally to the ratios of the priors.

The \textit{prediction} terms represent most of the complexity. They reflect the probability of a cell being currently on each possible state given the previous sensor readings and poses of the agent ${p(m_{t,i} \mid z_{1:t-1}, x_{1:t-1})}$.
Even if there existed an accurate prediction model, ${p(m_{t,i} \mid m_{t-1})}$, the state of the environment at the previous time step, $m_{t-1}$, is still uncertain, as it can only be estimated based on the agent's previous locations and sensor readings, $p(m_{t-1} \mid z_{1:t-1}, x_{1:t-1})$.
The computation of this term involves marginalizing over all the possible maps at the previous time step \eqref{eq:derivation2.1}, which grows exponentially with the size of the map and becomes unfeasible to compute. Section~\ref{sec:assumptions} introduces the main assumptions under which this term can be simplified. Section~\ref{sec:my_model} describes how, under those assumptions, we obtain an analytical solution.

\subsection{Random Transitions}
\label{sec:assumptions}

Let $\mathcal{T}_{j,i}$ denote the event that the content of cell $j$ transitions to cell $i$ between two consecutive time steps. Given $\mathcal{T}_{j,i}$, the state of $m_{t,i}$ after a transition is defined as
\begin{align} \label{eq:assumption1.1}
\vspace{-1mm}
    p(m^d_{t,i} \mid \mathcal{T}_{j,i}, m_{t-1}) =& m^d_{t-1,j},\\
    p(m^s_{t,i} \mid \mathcal{T}_{j,i}, m_{t-1}) =& m^s_{t-1,j},
\end{align}
i.e., if the event $\mathcal{T}_{j,i}$ occurs, the content of cell $j$ at time $t-1$ deterministically moves into cell $i$ at time $t$.
To prevent static obstacles from being able to move, and dynamic obstacles from being able to move to a static cell (depicted as red arrows in Fig.~\ref{fig:intro}), the probability of transition is assumed to depend on the previous state of the map, $m_{t-1}$:
\begin{equation} \label{eq:transitions}
    p(\mathcal{T}_{j,i} \mid m_{t-1}) = \begin{cases}

    \tau_{j,i} \overline{m^s_i} \ \overline{m^s_j},

    & \text{if } j \neq i \\

    m^s_i + \overline{m^s_i} \bigl( \tau_{i,i} + \sum_{k \neq i} \bigl[ \tau_{i,k} m^s_k \bigr] \bigr),

    & \text{if }j=i

    \end{cases}
\end{equation}

\setcounter{equation}{13}
\begin{table*}[b]
\centering
\begin{minipage}{\textwidth}
\hrule

\begin{gather}
    p(m^d_{t,i} \mid z_{1:t-1},x_{1:t-1}) = \sum_{m_{t-1}}\Bigl[ p(m^d_{t,i} \mid m_{t-1}) p(m_{t-1} \mid z_{1:t-1},x_{1:t-1}) \Bigr] \label{eq:derivation2.1}\\
    =
    \sum_{m_{t-1}}\Bigl[ \textcolor{blue}{\Bigl( p(m^d_{t-1,i} \mid m_{t-1}) \Bigl(\tau_{i,i} + \sum_{j \neq i} \Bigl[ \tau_{i,j} m^s_j \Bigr] \Bigr)
    +
    p(\overline{m^s_i} \mid m_{t-1}) \sum_{j \neq i} \Bigl[ p(m^d_{t-1,j} \mid m_{t-1}) \tau_{j,i} \Bigr] \Bigr)} p(m_{t-1} \mid z_{1:t-1},x_{1:t-1}) \Bigr] \label{eq:derivation2.2}\\
    = 
    \textcolor{blue}{\Bigl(\tau_{i,i} + \sum_{j \neq i} \Bigl[ \tau_{i,j} m^s_j \Bigr] \Bigr)} \! \! \sum_{m_{t-1}}\Bigl[ p(m^d_{t-1,i} \mid m_{t-1}) \textcolor{blue}{p(m_{t-1} \mid z_{1:t-1},x_{1:t-1})} \Bigr]
    +
    \sum_{m_{t-1}} \sum_{j \neq i} \Bigl[ \textcolor{blue}{p(\overline{m^s_i} \mid m_{t-1})} p(m^d_{t-1,j} \mid m_{t-1}) \tau_{j,i} \textcolor{blue}{p(m_{t-1} \mid z_{1:t-1},x_{1:t-1})} \Bigr] \label{eq:derivation2.3}\\
    = 
    \Bigl(\tau_{i,i} + \sum_{j \neq i} \Bigl[ \tau_{i,j} m^s_j \Bigr] \Bigr) \textcolor{blue}{p(m^d_{t-1,i} \mid z_{1:t-1},x_{1:t-1})}
    +
    \textcolor{blue}{\sum_{j \neq i} \Bigl[ \tau_{j,i} \sum_{m_{t-1}}} \bigl[ p(\overline{m^s_i} \mid m_{t-1})p(m^d_{t-1,j} \mid m_{t-1}) p(m_{t-1} \mid z_{1:t-1},x_{1:t-1}) \bigr] \Bigr] \label{eq:derivation2.4}\\
    = 
    \Bigl(\tau_{i,i} + \sum_{j \neq i} \Bigl[ \tau_{i,j} m^s_j \Bigr] \Bigl) p(m^d_{t-1,i} \mid z_{1:t-1},x_{1:t-1})
    +
    \textcolor{blue}{p(\overline{m^s_i} \mid m_{t-1})} \sum_{j \neq i} \Bigl[ \tau_{j,i} \textcolor{blue}{p(m^d_{t-1,j} \mid z_{1:t-1},x_{1:t-1})} \Bigr] \label{eq:derivation2.5}
\end{gather}

\medskip
\end{minipage}
\end{table*}

The initial transition probability, $\tau$, can be seen as the probability of a transition between two cells if there were no static cells around. In the presence of static cells, ${p(\mathcal{T}_{j,i} \mid m_{t-1})}$ is modified in \eqref{eq:transitions} as follows:

\begin{enumerate}
    \item The probability of a transition between two different cells is equal to 0 if any of the two cells are static.
    \item The probability of staying in the same cell, $p(\mathcal{T}_{i,i})$, becomes $1$ if the cell is static.
    \item The probability of staying in the same cell, $p(\mathcal{T}_{i,i})$, is also increased according to the number of other transitions that are not possible due to static cells in the neighborhood, making a dynamic obstacle surrounded by static cells more likely to stay in the same position.
\end{enumerate}

The prediction models $p(m_{t,i} \mid m_{t-1})$ can now be obtained by marginalizing over all possible $\mathcal{T}_{j,i}$:
\setcounter{equation}{11}
\begin{equation} \label{eq:pred.model_static}
    p(m^s_{t,i} \mid m_{t-1}) = \sum_j \Bigl[ m^s_{t-1,j}p(\mathcal{T}_{j,i} \mid m_{t-1}) \Bigr] = m^s_{t-1,i}
\end{equation}
\begin{multline} \label{eq:pred.model_dynamic}
\vspace{-2mm}
    p(m^d_{t,i} \mid m_{t-1}) = \sum_j \Bigl[ m^d_{t-1,j}p(\mathcal{T}_{j,i} \mid m_{t-1}) \Bigr]\\
    = m^d_{t-1,i}p(\mathcal{T}_{i,i} \mid m_{t-1}) + \sum_{j \neq i} \Bigl[ m^d_{t-1,j} p(\mathcal{T}_{j,i} \mid m_{t-1}) \Bigr]\\
    = m^d_{t-1,i} \Bigl( \tau_{i,i} + \sum_{k \neq i} \bigl[ \tau_{i,k}m^s_k \bigr] \Bigr) + \overline{m^s_i} \sum_{j \neq i} \Bigl[ m^d_{t-1,j} \tau_{j,i} \Bigr].
\end{multline}

\subsection{The Predictions}
\label{sec:my_model}
After deriving the prediction model ${p(m_{t,i} \mid m_{t-1})}$ based on random transitions \eqref{eq:pred.model_static}-\eqref{eq:pred.model_dynamic}, the aim now is to obtain a prediction based on previous observations ${p(m_{t,i} \mid z_{1:t-1}, x_{1:t-1})}$, that does not require marginalizing over all possible maps.

Since static cells are predicted to stay in the same place \eqref{eq:pred.model_static}, the static prediction is equal to the previous probability $p(m^s_t \mid z_{1:t-1}, x_{1:t-1}) = p(m^s_{t-1} \mid z_{1:t-1}, x_{1:t-1})$.

When including the prediction model assuming random transitions ~\eqref{eq:pred.model_dynamic}, the prediction for the dynamic cells can be simplified as shown in \eqref{eq:derivation2.1}-\eqref{eq:derivation2.5}.
It is important to highlight that these predictions do not require marginalizing over all the $3^N$ possible maps at the previous time-step, $m_{t-1}$, making the computation feasible.

\subsection{Transition Probability Independent of Absolute Location}
\label{sec:convolution}

So far, for simplicity of the notation, cells have been referenced by one index, e.g., $i$ or $j$. In this section, they are referenced by their $x$ and $y$ components. In some cases, the initial probability of transition between two cells does not depend on the absolute position of those cells, but only on the relative position between them, such that $T[\Delta x, \Delta y] = T[x_i-x_j,y_i-y_j] = \tau_{j,i}$. In those cases, the prediction can be simplified as a set of operations involving a couple of 2D convolutions of the static and dynamic maps.

We define $M^s$ and $M^d$ as functions storing the previously estimated static and dynamic map, respectively:
\setcounter{equation}{18}
\begin{align}
    M^s[x_i,y_i] &= p(m^s_i \mid z_{1:t-1},x_{1:t-1}),\\
    M^d[x_i,y_i] &= p(m^d_{t-1,i} \mid z_{1:t-1},x_{1:t-1}).
\end{align}
We also define the kernel $K$ as the initial probability of transition for different cells, and $K'$, its flipped version:
\begin{align}
    K[\Delta x,\Delta y] &= T[\Delta x,\Delta y] \llbracket \Delta x, \Delta j \neq 0 \rrbracket,\\
    K'[\Delta x,\Delta y] &= T[-\Delta x,-\Delta y] \llbracket \Delta x, \Delta j \neq 0 \rrbracket.
\end{align}

In this case, the prediction term from \eqref{eq:derivation2.5} can be simplified to the following expression, involving a pair of discrete 2D convolutions, denoted by $\ast \ast$:
\begin{equation} \label{eq:convolution_model}
    \begin{aligned}
        p(m^d_{t,i} \mid z_{1:t}, x_{1:t}) = M^d[x_i,y_i] \left( \tau_0 + (M^s \ast \vphantom{h} \! \! \ast K')[x_i,y_i] \right)
        \\ +
        (1-M^s[x_i,y_i])(M^d \ast \vphantom{h} \! \! \ast K)[x_i,y_i]
    \end{aligned}
\end{equation}
where $\tau_0$ denotes $T[0,0]$.

With this model, the filter works by sequentially smoothing out the previous belief according to where dynamic obstacles can move within the static map and, subsequently, integrating the new observations into the smoothed out belief.

\subsection{Transition Distribution}
\label{sec:implementation}

The previous subsections of the method assume that the initial distribution of the random transitions between cells, $p(\tau_{j,i})$, is known. This distribution could be learned, similarly to \citep{kucner_conditional_2013} or \citep{wang_modeling_2014}, or it could be derived from a model. In this work, the latter approach has been chosen, since learning a model requires application-specific data.

We assume a maximum speed at which an obstacle can move, $v_\mathrm{max}$, which, together with the time step $\Delta t$, limits the distance $d_\mathrm{max}=v_\mathrm{max}/\Delta t$ that the content of a cell can move in one time step. In our experiments, we assume a uniform distribution, i.e., it is equally likely that the content of a cell moves to any of the cells within the distance $d_\mathrm{max}$:

\begin{equation}\label{eq:T}
    T[\Delta x,\Delta y] =
    \begin{cases}
    1/n & \sqrt{{\Delta x}^2+{\Delta y}^2} \le d_\mathrm{max}\\
    0 & \text{otherwise} ,
    \end{cases}
\end{equation}
where $n$ is the number of inputs, $(\Delta x,\Delta y)$, that satisfy the condition, $\sqrt{{\Delta x}^2+{\Delta y}^2} \le d_\mathrm{max}$.

\subsection{Simultaneous Localization and Mapping}
\label{sec:SLAM}

The proposed method assumes the poses to be known, similar to the Occupancy Grid Map. This assumption can be relaxed by combining the proposed Transitional Grid Maps with an existing SLAM approach. As an example, the scan matching approach proposed in \cite{hess_real-time_2016} has been used. The task of finding the most likely scan pose is formulated as a nonlinear least squares problem:

\begin{equation}\label{eq:SLAM}
    x_t = \operatorname*{arg\,min}_x \sum_{k=1}^K \left( 1 - M^s_\text{smooth} [T_x z_t^k]\right)^2,
\end{equation}
where $z_t^k$ denotes each of the scan points on the measurement $z_t$, $T_x$ transforms $z_t^k$ from the scan frame to the map frame according to the pose $x$, and $M^s_\text{smooth}$ denotes a smooth version of the probability values in the static map $M^s$.

The difference with the original approach is that instead of matching the new scan with a regular grid map, it is matched with the static part of the TGM. This can help the SLAM in highly dynamic environments, as shown in Sec.~\ref{sec:exp_3}.
\section{Experiments}
We study the behavior of the proposed TGM in two experiments and compare the results against two baselines.
As the first baseline for the evaluation, we use an Occupancy Grid Map (OGM) in combination with the same SLAM approach, as originally presented in \cite{hess_real-time_2016}.
As the second baseline, we include a clamped Occupancy Grid Map (c-OGM), where the confidence about each cell has a saturation limit, similarly to \citep{kraetzschmar_probabilistic_2004, hornung2013octomap}.
For the TGM, the implementation follows \eqref{eq:convolution_model} in combination with the grid map-based SLAM approach presented in \cite{hess_real-time_2016}, as formulated in \eqref{eq:SLAM}.
Table~\ref{tab:parameters} shows the values used for the main parameters of the proposed method and the baselines. Under these settings, the current Python implementation of the TGM takes $\SI{0.1}{\s/update}$ and the SLAM $\SI{0.015}{\s/update}$, closely matching the frequency of the sensor, $\SI{10}{\hertz}$.
For large maps, such as the one in Figure~\ref{fig:mapping_3}, we only update the area close to the ego vehicle.

\begin{figure}[b]
    \centering
    \includegraphics[width=1\linewidth]{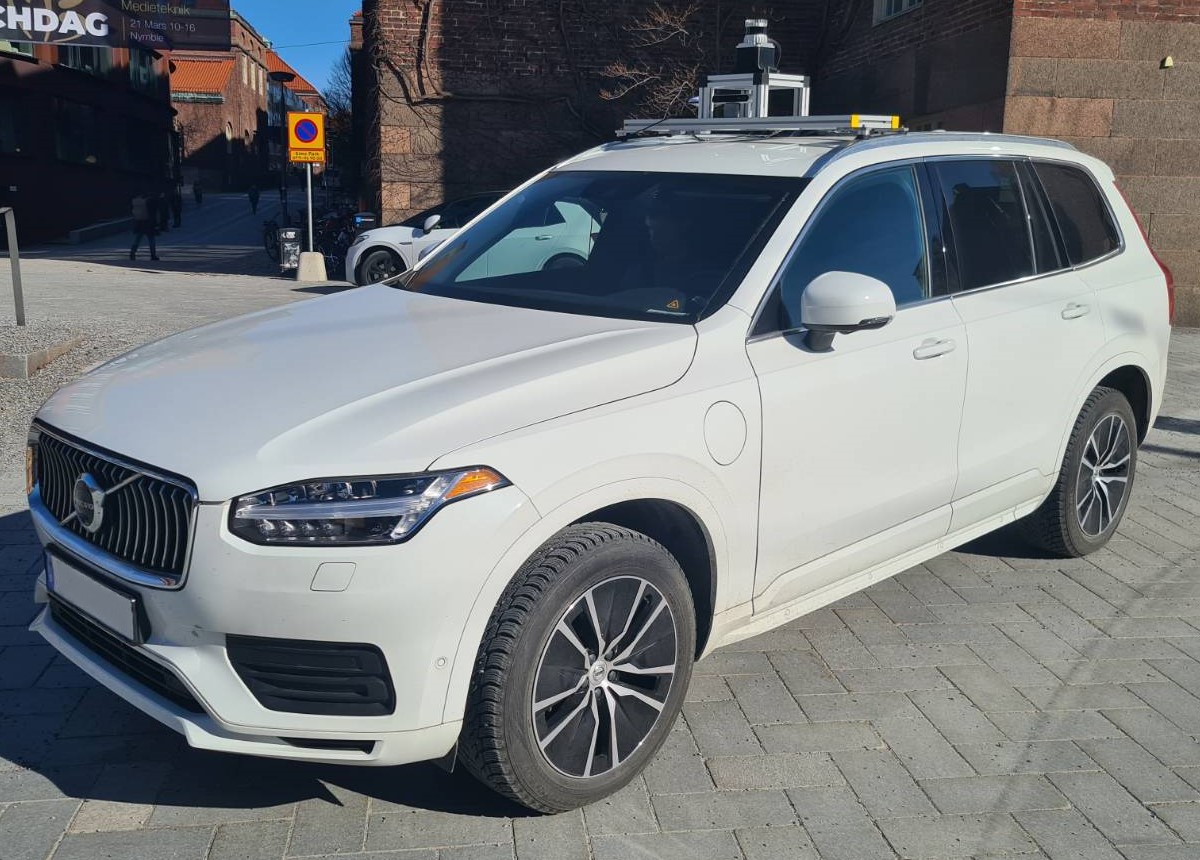}
    \vspace{-0.5cm}
    \caption{Volvo XC90 used for the data collection with sensors at the top.}
    \label{fig:volvo}
\end{figure}

In section~\ref{sec:exp_1}, we evaluate the quality of the maps produced while driving in an urban environment. Section~\ref{sec:exp_3} presents a particularly challenging scenario for SLAM and evaluates how the proposed approach can help mitigate these challenges.
All the experimental data has been recorded using a Volvo XC90 equipped with an Ouster OS1-32 lidar, as seen in Fig.~\ref{fig:volvo}. The vehicle is also equipped with a ZED camera and an RTK-SSR Receiver, not used by the algorithm.

\begin{figure}[t]
    \centering
    \includegraphics[width=0.7\linewidth]{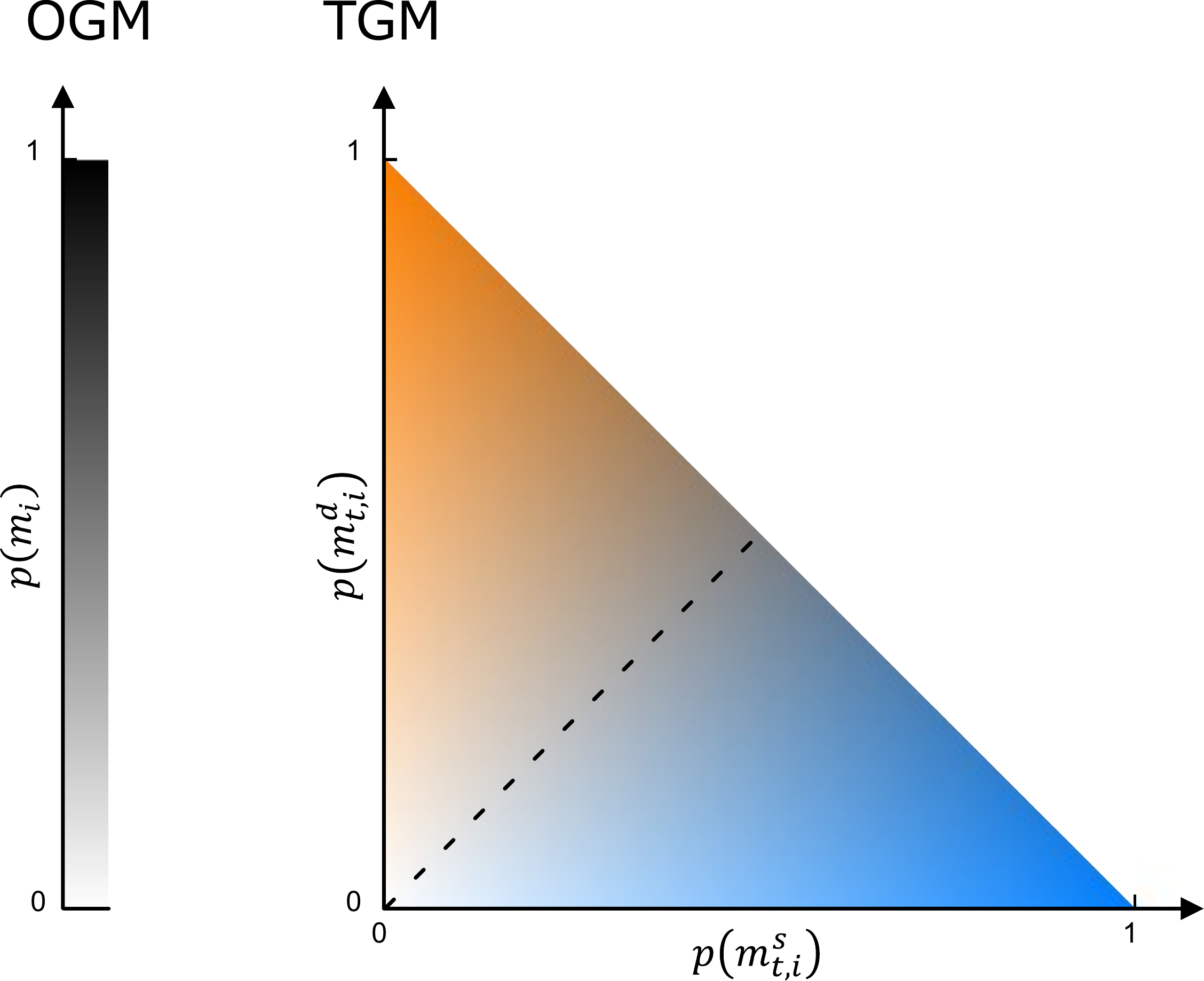}
    \caption{Color map used to represent the current belief about each cell for the baselines (left) and for TGMs (right).}
    \label{fig:plots}
\end{figure}

Cells are colored based on the current belief about their state. For the baseline, a gray scale is used to represent the probability of occupancy. For the TGMs, the belief about each cell is described by the belief about the cell being static, $p(m^s_{t,i})$, and the belief about the cell being dynamic, $p(m^d_{t,i})$. Each of them can be any value between 0 and 1, while the sum of both must be smaller or equal to 1. The remainder is the probability of the cell being free, $p(m^f_{t,i})$. Figure~\ref{fig:plots} shows the color map used. Free cells are depicted in white, while blue and orange are used for static and dynamic respectively. Using complementary colors ensures that when $p(m^s_{t,i}) = p(m^d_{t,i})$, e.g., for cells that have not been observed, the resulting color is on a gray scale (dashed line in Fig.~\ref{fig:plots}).

\begin{table}[htbp]
\centering
\caption{Experimental parameters}
\label{tab:parameters}
\newcommand{\notapplicable}{}
\setlength{\tabcolsep}{4pt}
\begin{tabular}{@{}lcccccr@{}}
\toprule
 &  & \multicolumn{2}{c}{Priors} & \multicolumn{2}{c}{Saturation Limits} & \\ \cmidrule(lr){3-4} \cmidrule(lr){5-6}
Exp. & Setup & $p(m^s_{t,i})$ & $p(m^d_{t,i})$ & $p(m^s_{t,i})$ & $p(m^d_{t,i})$ & Range \\ \midrule
\multirow{3}{*}{A} & OGM & 0.5 & \notapplicable & 0 -- 1 & \phantom{0.05}\notapplicable\phantom{1} & 100 m \\ 
 & c-OGM & 0.5 & \notapplicable & 0.05 -- 0.95 & \phantom{0.05}\notapplicable\phantom{1} & 100 m \\
 & TGM & 0.3 & 0.3 & \phantom{0.0}0 -- 0.95 & 0.05 -- 1 & 100 m \\ \midrule
\multirow{3}{*}{B} & OGM & 0.5 & \notapplicable & 0 -- 1 & \phantom{0.05}\notapplicable\phantom{1} & 25 m \\ 
 & c-OGM & 0.5 & \notapplicable & 0.05 -- 0.95 & \phantom{0.05}\notapplicable\phantom{1} & 25 m \\ 
 & TGM & 0.3 & 0.3 & \phantom{0.0}0 -- 0.95 & 0.05 -- 1 & 25 m \\ \bottomrule
\end{tabular}
\end{table}

\subsection{Improved Static Mapping}
\label{sec:exp_1}
\begin{figure}[t]
    \centering
    \includegraphics[width=1\linewidth]{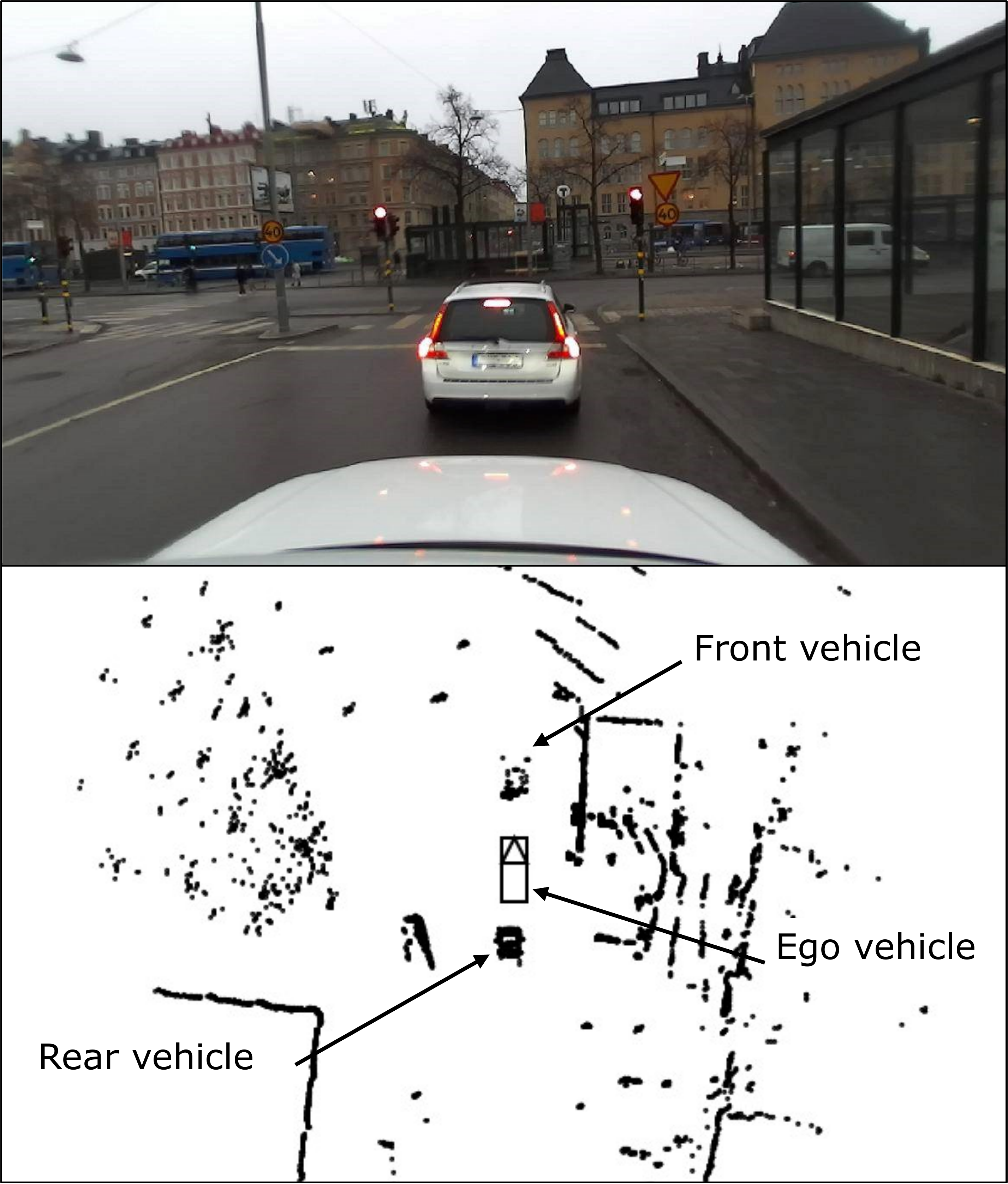}
    \caption{Snapshot of the scenario recorded for the first experiment. The image from the camera shows the ego vehicle stopped with a vehicle in front (top). The lidar scan shows the same vehicle at the front, another vehicle at the back and a few buildings (bottom).}
    \label{fig:mapping_1}
\end{figure}

We first compare the quality of the maps generated during a common driving situation. Figure~\ref{fig:mapping_1} shows a scene from the recorded scenario. The ego vehicle stops at a traffic light with one vehicle in front and another one behind. After some time, the traffic light turns green and all of them move.

\begin{figure}[h]
    \centering
    \includegraphics[width=1\linewidth]{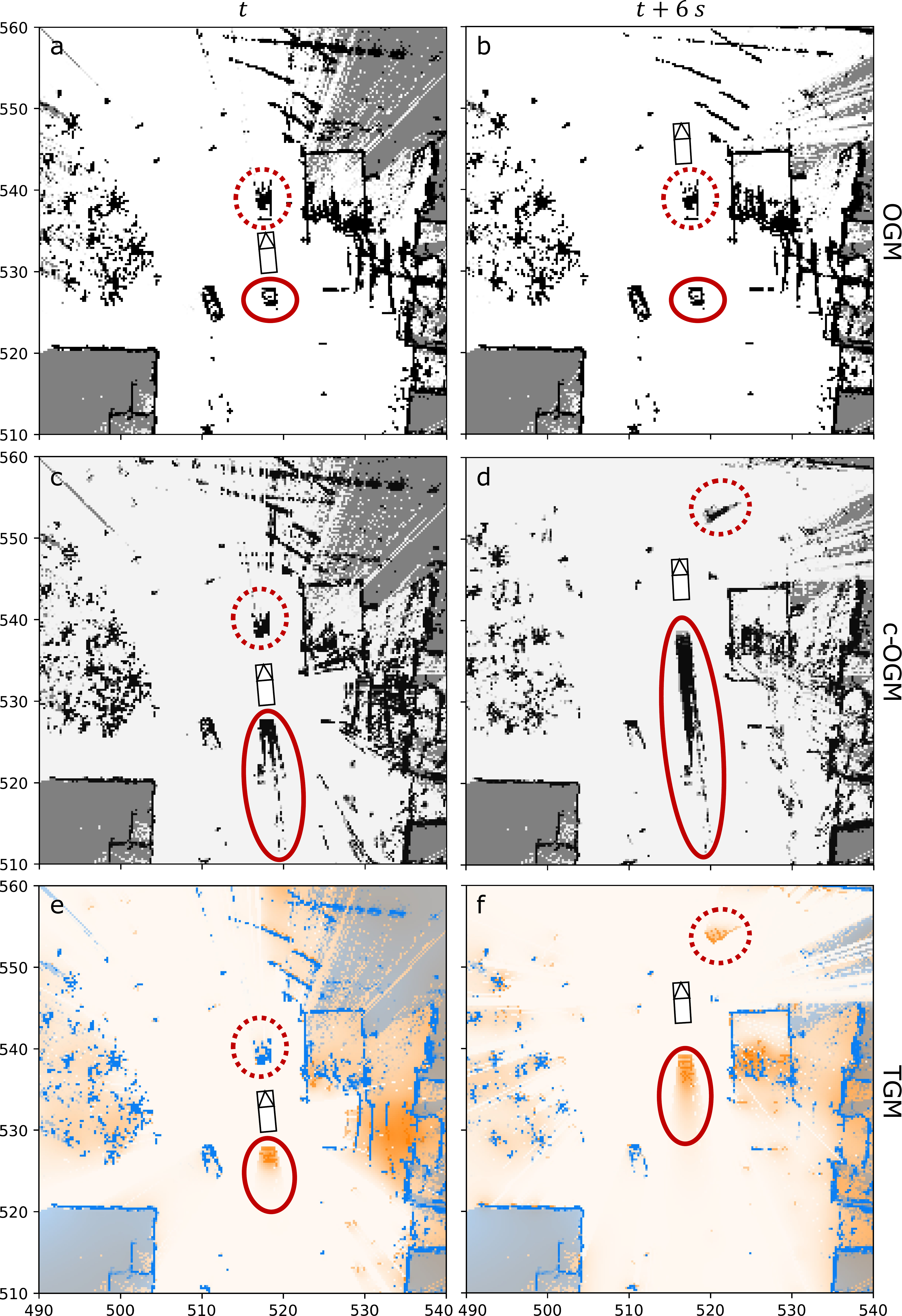}
    \vspace{-0.2cm}
    \caption{Results from the first experiment. The ego vehicle is shown with a rectangle. (a)-(b) The OGM becomes overconfident about the stopped cars at the traffic light (highlighted with circles), preventing them from being removed from the map when they move. (c)-(d) The c-OGM shifts the state of cells quicker due to the saturation limit, but produces traces along moving obstacles. (e)-(f) The TGM correctly classifies moving vehicles as dynamic (orange), preventing them from becoming part of the static map (blue).}
    \label{fig:mapping_2}
    \vspace{-0.2cm}
\end{figure}

Figures~\ref{fig:mapping_2}.a-b show the first baseline. Due to the lack of saturation limit, the OGM becomes very confident about the occupancy of the cells where the other vehicles are (Fig.~\ref{fig:mapping_2}.a). This has the result that when the vehicles move, their cells are not updated quickly enough, leaving their previous poses as part of the map (Fig.~\ref{fig:mapping_2}.b).

The second baseline also becomes quite confident about the occupancy of the cells populated by the other vehicles (Fig.~\ref{fig:mapping_2}.c). In contrast, due to the confidence being capped at the saturation limit, the cells can shift their belief when new contradicting measurements arrive. This comes at the cost of leaving traces in cells that are not observed free afterwards, like the ones caused by the vehicle behind (Fig.~\ref{fig:mapping_2}.d).

Figures~\ref{fig:mapping_2}.e-f show the results using the TGM. Initially, the vehicle in the front is perceived as static, since it was not observed moving before, while the vehicle in the back is correctly inferred as dynamic (Fig.~\ref{fig:mapping_2}.e). Afterwards, when the front vehicle moves, those cells quickly shift to dynamic, leaving no traces in the static map (Fig.~\ref{fig:mapping_2}.f).

\begin{figure}[H]
    \centering
    \includegraphics[width=0.95\linewidth]{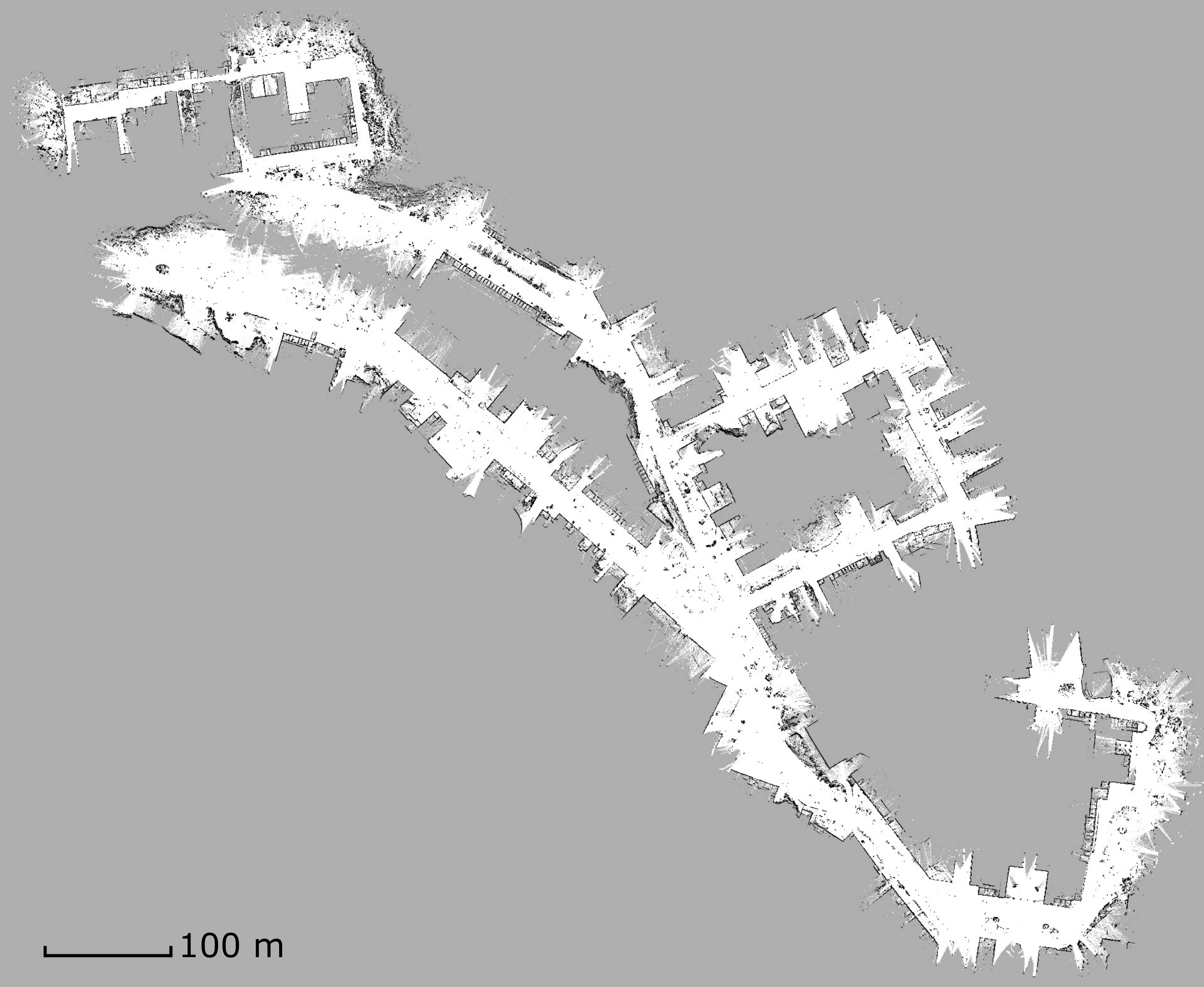}
    \caption{Example of the map obtained by SLAM using TGMs at the KTH campus. Only the static part of the map has been plotted, showing no signs of traces from dynamic obstacles.}
    \label{fig:mapping_3}
\end{figure}

Figure~\ref{fig:mapping_3} shows an example of the static map obtained using a TGM at the KTH campus, showing no signs of traces from other road users.

It should be highlighted that being able to produce an accurate map of an unknown environment and being able to update it continuously when new measurements are available is crucial for downstream tasks, such as planning.

\vspace{-1cm}

\setcounter{figure}{8}

\begin{figure*}[b]
    \centering
    \includegraphics[width=1\linewidth]{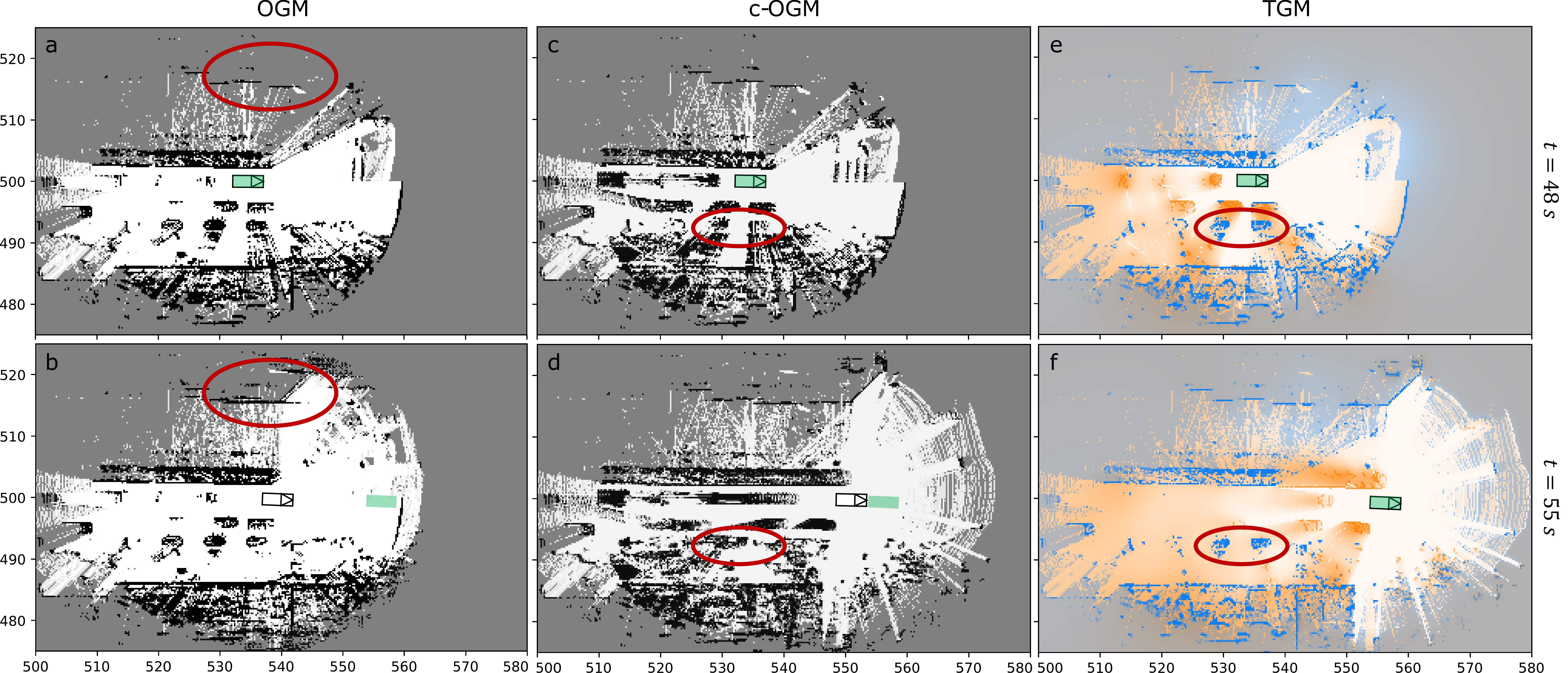}
    \caption{Results from the second experiment. The estimated pose of the ego vehicle is shown with an outlined rectangle, while the ground truth pose is shown with a green rectangle. (a)-(b) The OGM starts with the right pose but completely drifts from the ground truth. (c)-(d) The c-OGM has a smaller, but considerable drift, making some of the parked cars appear elongated. (e)-(f) The TGM has no drift, resulting in no apparent distortions of the map.}
    \label{fig:slam2}
\end{figure*}

\subsection{SLAM in Highly Dynamic Environments}
\label{sec:exp_3}
In the last experiment, we present a particularly challenging scenario for SLAM due to the high number of dynamic actors around the ego vehicle.

\setcounter{figure}{7}

\begin{figure}[H]
    \centering
    \includegraphics[width=1\linewidth]{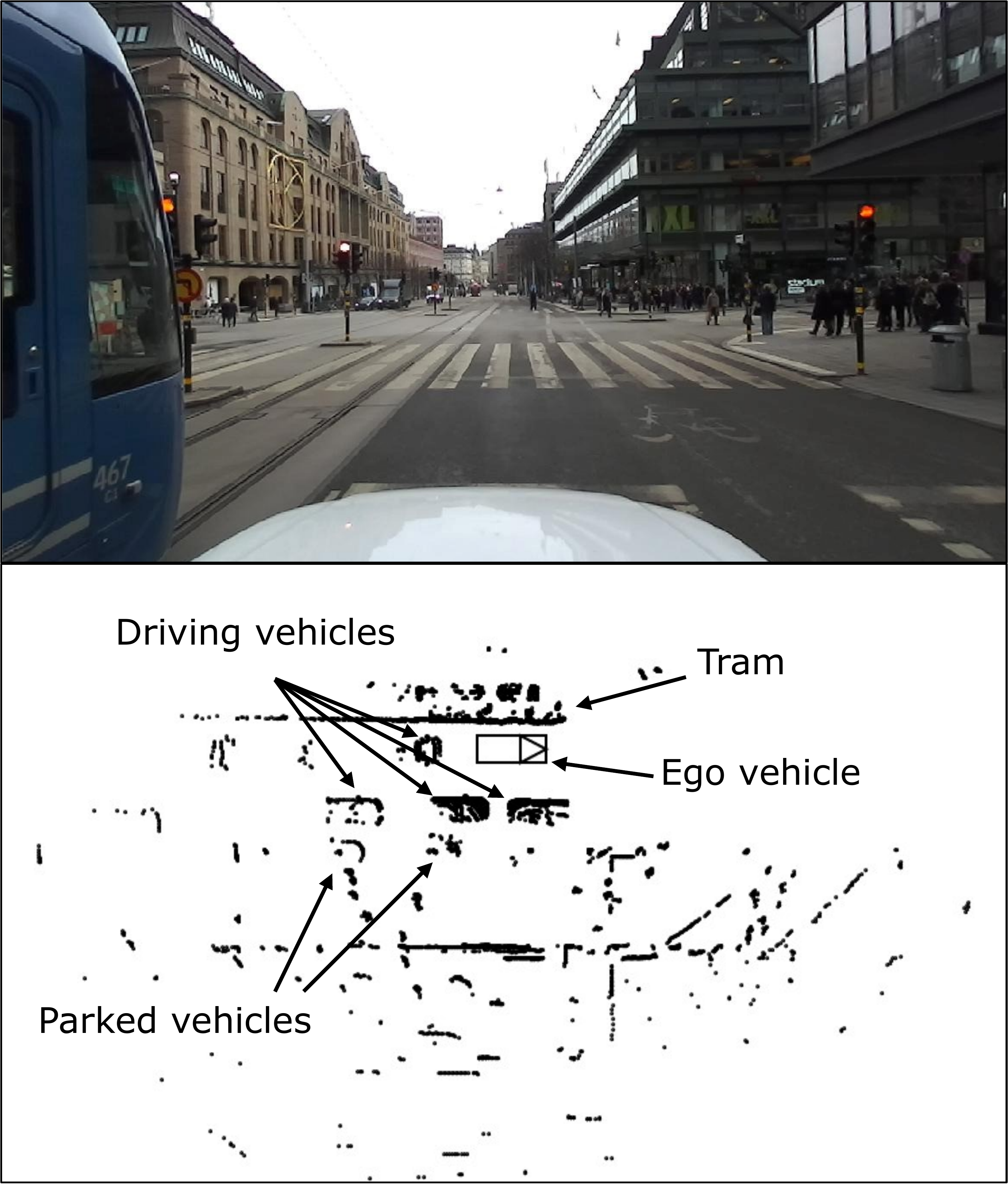}
    \vspace{-0.2cm}
    \caption{Scene from the scenario recorded for the second experiment. The image from the camera shows the ego vehicle stopped next to a tram (top). The lidar scan shows the tram, other vehicles, and some buildings (bottom).}
    \label{fig:slam1}
    \vspace{-0.2cm}
\end{figure}

Figure~\ref{fig:slam1} shows a scene from the recorded scenario. First, the ego vehicle stops at a traffic light, next to an already-stopped tram. Later, more vehicles arrive, stopping behind and to the side of the ego vehicle. Finally, after some time has passed, the traffic light turns green and all the vehicles start moving simultaneously.
To create a more challenging scenario for the SLAM algorithm, the range of the lidar sensor has been limited to 25 meters, preventing the algorithm from taking advantage of the high buildings present. The SLAM results without restricting the lidar's range have been used as ground truth.

Figure~\ref{fig:slam2}.a shows the OGM right before the traffic light turns green. The map contains the tram, the parked cars, and some features of the buildings in the south, but also the cars that have arrived after the ego vehicle since they have been observed long enough.
Figure~\ref{fig:slam2}.b shows the OGM shortly after the traffic light turns green. Due to the high number of dynamic elements and since all of them start moving in the same direction, the SLAM algorithm wrongly matches the new scans with the moving vehicles instead of with the static elements, leading to a wrong localization and erroneous update of the map. This can also be observed in Figure~\ref{fig:slam3}, where there is a strong mismatch between the pose estimated by the first baseline and the ground truth computed without restricting the lidar.

In Figure~\ref{fig:slam2}.c, the c-OGM shows all vehicles visible, similarly to the OGM with some extra traces.
In Figure~\ref{fig:slam2}.d, it can be observed that the result of the SLAM is slightly off from the ground truth, leading to some artifacts that can be seen around the parked vehicles and the buildings. The same small error can be observed in Figure~\ref{fig:slam3}, where the c-OGM has a constant offset with respect to the ground truth after the vehicles start moving.

Figure~\ref{fig:slam2}.e shows the TGM before the traffic light turns green. The tram, the parked vehicles, and some features from the buildings at the bottom are perceived as highly likely static (dark blue), as they have been consistently observed occupied. In contrast, the vehicles behind and to the right of the ego vehicle are perceived as probably dynamic (dark orange). Some areas outside the current field of view are marked with a light shade of orange, meaning a probability of being dynamic close to the prior but not likely to be static since they have been observed free before. It is important to highlight that this behavior is not hard-coded, but emerges from the random transitions assumed for dynamic cells.
Figure~\ref{fig:slam2}.f shows the state of the TGM shortly after the traffic light turns green and the vehicles have started moving. Despite the high number of dynamic elements, since the SLAM only tries to match new lidar measurements with the static part of the TGM (as described in Sec.~\ref{sec:SLAM}), the localization works properly as it can be observed by the continuity of the buildings and parked cars, and also by the overlap in Figure~\ref{fig:slam3}.

\setcounter{figure}{9}

\begin{figure}[h]
    \centering
    \includegraphics[width=1\linewidth]{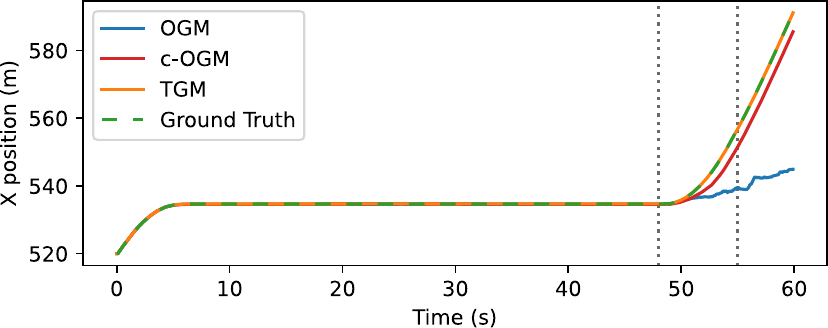}
    \caption{X component of the estimated pose when performing SLAM using the baselines and the TGM. The dashed vertical lines indicate the timestamps visualized in Figure~\ref{fig:slam2}.}
    \label{fig:slam3}
\end{figure}
\section{Conclusions and Future Work}

This work formalizes the problem of simultaneously inferring the static and dynamic parts of an unknown environment and introduces Transitional Grid Maps (TGMs) as an analytical solution.
Using real data, we demonstrate how this approach produces better maps by keeping track of both static and dynamic elements and, as a side effect, can improve existing SLAM algorithms.

These benefits could also help other downstream tasks, such as planning, which should be studied in the future.
Additionally, the method could be combined with a tracker to provide velocity estimates to the dynamic cells. Finally, classification information obtained from other sensors, such as cameras, could be incorporated into the grid, allowing for different transition models per class.

\bibliographystyle{IEEEtran}
\bibliography{references}

\begin{thebibliography}{10}
\providecommand{\url}[1]{#1}
\csname url@samestyle\endcsname
\providecommand{\newblock}{\relax}
\providecommand{\bibinfo}[2]{#2}
\providecommand{\BIBentrySTDinterwordspacing}{\spaceskip=0pt\relax}
\providecommand{\BIBentryALTinterwordstretchfactor}{4}
\providecommand{\BIBentryALTinterwordspacing}{\spaceskip=\fontdimen2\font plus
\BIBentryALTinterwordstretchfactor\fontdimen3\font minus \fontdimen4\font\relax}
\providecommand{\BIBforeignlanguage}[2]{{%
\expandafter\ifx\csname l@#1\endcsname\relax
\typeout{** WARNING: IEEEtran.bst: No hyphenation pattern has been}%
\typeout{** loaded for the language `#1'. Using the pattern for}%
\typeout{** the default language instead.}%
\else
\language=\csname l@#1\endcsname
\fi
#2}}
\providecommand{\BIBdecl}{\relax}
\BIBdecl

\bibitem{elfes_using_1989}
A.~Elfes, ``Using occupancy grids for mobile robot perception and navigation,'' \emph{Computer}, vol.~22, no.~6, pp. 46--57, 1989.

\bibitem{moravec_high_1985}
H.~Moravec and A.~Elfes, ``High resolution maps from wide angle sonar,'' in \emph{Proceedings of the IEEE International Conference on Robotics and Automation}, vol.~2, 1985, pp. 116--121.

\bibitem{moravec_sensor_1989}
H.~P. Moravec, ``Sensor fusion in certainty grids for mobile robots,'' in \emph{Sensor Devices and Systems for Robotics}.\hskip 1em plus 0.5em minus 0.4em\relax Springer, 1989, pp. 253--276.

\bibitem{wang2007simultaneous}
C.-C. Wang, C.~Thorpe, S.~Thrun, M.~Hebert, and H.~Durrant-Whyte, ``Simultaneous localization, mapping and moving object tracking,'' \emph{The International Journal of Robotics Research}, vol.~26, no.~9, pp. 889--916, 2007.

\bibitem{coue_using_2003}
C.~Coue, T.~Fraichard, P.~Bessiere, and E.~Mazer, ``Using {Bayesian} programming for multi-sensor multi-target tracking in automotive applications,'' in \emph{Proceedings of the IEEE International Conference on Robotics and Automation}, vol.~2, 2003, pp. 2104--2109.

\bibitem{coue_bayesian_2006}
C.~Coué, C.~Pradalier, C.~Laugier, T.~Fraichard, and P.~Bessière, ``Bayesian occupancy filtering for multitarget tracking: An automotive application,'' \emph{The International Journal of Robotics Research}, vol.~25, no.~1, pp. 19--30, 2006.

\bibitem{cooper_computational_1990}
G.~F. Cooper, ``The computational complexity of probabilistic inference using bayesian belief networks,'' \emph{Artificial Intelligence}, vol.~42, no. 2-3, pp. 393--405, 1990.

\bibitem{yguel_update_2008}
M.~Yguel, O.~Aycard, and C.~Laugier, ``Update policy of dense maps: Efficient algorithms and sparse representation,'' in \emph{Field and Service Robotics}.\hskip 1em plus 0.5em minus 0.4em\relax Springer, 2008, pp. 23--33.

\bibitem{kraetzschmar_probabilistic_2004}
G.~K. Kraetzschmar, G.~P. Gassull, and K.~Uhl, ``Probabilistic quadtrees for variable-resolution mapping of large environments,'' \emph{IFAC Proceedings Volumes}, vol.~37, no.~8, pp. 675--680, 2004.

\bibitem{hornung2013octomap}
A.~Hornung, K.~M. Wurm, M.~Bennewitz, C.~Stachniss, and W.~Burgard, ``{OctoMap}: An efficient probabilistic {3D} mapping framework based on octrees,'' \emph{Autonomous Robots}, vol.~34, no.~3, pp. 189--206, 2013.

\bibitem{meyer-delius_occupancy_2012}
D.~Meyer-Delius, M.~Beinhofer, and W.~Burgard, ``Occupancy grid models for robot mapping in changing environments,'' in \emph{Proceedings of the AAAI Conference on Artificial Intelligence}, vol.~26, no.~1, 2012, pp. 2024--2030.

\bibitem{wang_modeling_2014}
Z.~Wang, R.~Ambrus, P.~Jensfelt, and J.~Folkesson, ``Modeling motion patterns of dynamic objects by {IOHMM},'' in \emph{Proceedings of the IEEE/RSJ International Conference on Intelligent Robots and Systems}, 2014, pp. 1832--1838.

\bibitem{steyer_object_2017}
S.~Steyer, G.~Tanzmeister, and D.~Wollherr, ``Object tracking based on evidential dynamic occupancy grids in urban environments,'' in \emph{IEEE Intelligent Vehicles Symposium (IV)}, 2017, pp. 1064--1070.

\bibitem{steyer_grid-based_2018}
------, ``Grid-based environment estimation using evidential mapping and particle tracking,'' \emph{IEEE Transactions on Intelligent Vehicles}, vol.~3, no.~3, pp. 384--396, 2018.

\bibitem{tanzmeister_evidential_2017}
G.~Tanzmeister and D.~Wollherr, ``Evidential grid-based tracking and mapping,'' \emph{IEEE Transactions on Intelligent Transportation Systems}, vol.~18, no.~6, pp. 1454--1467, 2017.

\bibitem{nuss_random_2018}
D.~Nuss, S.~Reuter, M.~Thom, T.~Yuan, G.~Krehl, M.~Maile, A.~Gern, and K.~Dietmayer, ``A random finite set approach for dynamic occupancy grid maps with real-time application,'' \emph{The International Journal of Robotics Research}, vol.~37, no.~8, pp. 841--866, 2018.

\bibitem{dempster_generalization_1968}
A.~P. Dempster, ``A generalization of {Bayesian} inference,'' \emph{Journal of the Royal Statistical Society. Series B (Methodological)}, vol.~30, no.~2, pp. 205--247, 1968.

\bibitem{shafer_mathematical_1976}
G.~Shafer, \emph{A Mathematical Theory of Evidence}.\hskip 1em plus 0.5em minus 0.4em\relax Princeton University Press, 1976.

\bibitem{rexin_modeling_2017}
N.~Rexin, D.~Nuss, S.~Reuter, and K.~Dietmayer, ``Modeling occluded areas in dynamic grid maps,'' in \emph{Proceedings of the IEEE International Conference on Information Fusion (FUSION)}, Jul. 2017, pp. 1--6.

\bibitem{dequaire_deep_2018}
\BIBentryALTinterwordspacing
J.~Dequaire, P.~Ondrúška, D.~Rao, D.~Wang, and I.~Posner, ``Deep tracking in the wild: {End}-to-end tracking using recurrent neural networks,'' \emph{The International Journal of Robotics Research}, vol.~37, no. 4-5, pp. 492--512, Apr. 2018. [Online]. Available: \url{http://journals.sagepub.com/doi/10.1177/0278364917710543}
\BIBentrySTDinterwordspacing

\bibitem{schreiber_dynamic_2021}
M.~Schreiber, V.~Belagiannis, C.~Gläser, and K.~Dietmayer, ``Dynamic {Occupancy} {Grid} {Mapping} with {Recurrent} {Neural} {Networks},'' in \emph{2021 {IEEE} {International} {Conference} on {Robotics} and {Automation} ({ICRA})}, May 2021, pp. 6717--6724.

\bibitem{kucner_conditional_2013}
T.~Kucner, J.~Saarinen, M.~Magnusson, and A.~J. Lilienthal, ``Conditional transition maps: Learning motion patterns in dynamic environments,'' in \emph{Proceedings of the IEEE/RSJ International Conference on Intelligent Robots and Systems}, 2013, pp. 1196--1201.

\bibitem{hess_real-time_2016}
W.~Hess, D.~Kohler, H.~Rapp, and D.~Andor, ``Real-time loop closure in {2D} {LIDAR} {SLAM},'' in \emph{2016 {IEEE} {International} {Conference} on {Robotics} and {Automation} ({ICRA})}, May 2016, pp. 1271--1278.

\end{thebibliography}

\balance

\end{document}